\documentclass[letterpaper]{article} 
\usepackage{aaai24}  
\usepackage{times}  
\usepackage{helvet}  
\usepackage{courier}  
\usepackage[hyphens]{url}  
\usepackage{graphicx} 
\usepackage{natbib}  
\usepackage{caption} 
\usepackage{algorithm,algpseudocode}
\usepackage{multirow}
\usepackage{subfigure}
\usepackage{makecell}
\usepackage{color}
\usepackage{amsmath}
\usepackage{amssymb}
\usepackage{gensymb}

\usepackage{newfloat}
\usepackage{listings}
\DeclareCaptionStyle{ruled}{labelfont=normalfont,labelsep=colon,strut=off} 
\floatstyle{ruled}
\newfloat{listing}{tb}{lst}{}
\floatname{listing}{Listing}
\title{MetaDiff: Meta-Learning with Conditional Diffusion for Few-Shot Learning}
\author{
    Baoquan Zhang\textsuperscript{\rm 1},
    Chuyao Luo\textsuperscript{\rm 1},
    Demin Yu\textsuperscript{\rm 1},
    Xutao Li\textsuperscript{\rm 1}, 
    Huiwei Lin\textsuperscript{\rm 1}, 
    Yunming Ye\textsuperscript{\rm 1}\thanks{Corresponding author.},
    Bowen Zhang\textsuperscript{\rm 2}
}
\affiliations{
    \textsuperscript{\rm 1}Harbin Institute of Technology, Shenzhen\\
    \textsuperscript{\rm 2}Shenzhen Technology University \\
   baoquanzhang@hit.edu.cn, luochuyao.dalian@gmail.com, deminyu98@gmail.com, lixutao@hit.edu.cn, linhuiwei@stu.hit.edu.cn, yeyunming@hit.edu.cn, zhang\_bo\_wen@foxmail.com
}

\usepackage{bibentry}

\begin{document}

\maketitle

\begin{abstract}
Equipping a deep model the ability of few-shot learning (FSL) is a core challenge for artificial intelligence. Gradient-based meta-learning effectively addresses the challenge by learning how to learn novel tasks. Its key idea is learning a deep model in a bi-level optimization manner, where the outer-loop process learns a shared gradient descent algorithm (called meta-optimizer), while the inner-loop process leverages it to optimize a task-specific base learner with few examples. Although these methods have shown superior performance on FSL, the outer-loop process requires calculating second-order derivatives along the inner-loop path, which imposes considerable memory burdens and the risk of vanishing gradients. This degrades meta-learning performance. Inspired by recent diffusion models, we find that the inner-loop gradient descent process can be viewed as a reverse process (\emph{i.e.}, denoising) of diffusion where the target of denoising is the weight of base learner but origin data. Based on this fact, we propose to model the gradient descent algorithm as a diffusion model and then present a novel conditional diffusion-based meta-learning, called MetaDiff, that effectively models the optimization process of base learner weights from Gaussian initialization to target weights in a denoising manner. Thanks to the training efficiency of diffusion models, our MetaDiff does not need to differentiate through the inner-loop path such that the memory burdens and the risk of vanishing gradients can be effectively alleviated for improving FSL. Experimental results show that our MetaDiff outperforms state-of-the-art gradient-based meta-learning family on FSL tasks.
\end{abstract}

\section{Introduction}
\label{sec:intro}
With a large number of labeled data, deep learning techniques have shown superior performance and made breakthrough on various tasks. However, collecting such much data may be impractical or very difficult on some applications such as drug screening \cite{Altae-TranRPP16} and cold-start recommendation \cite{VartakTMBL17}. Inspired by the fast learning abaility of humans, \emph{i.e.}, humans can quickly learn a new concept or task from only very few examples, few-shot learning (FSL) has been proposed and has gained wide attention. It aims to learn transferable knowledge from data-ubundant base classes and then assist novel class prediction with few labeled examples \cite{WangYKN20}.

\begin{figure}[t!]
	\centering
	\subfigure[Workflow of gradient descent algorithm]{
		\includegraphics[width=1.0\columnwidth]{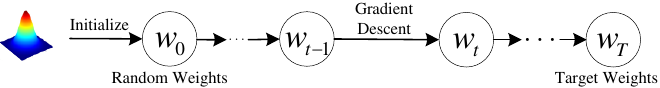}
		\label{fig1a}}
	\subfigure[Workflow of diffusion models]{
		\includegraphics[width=1.0\columnwidth]{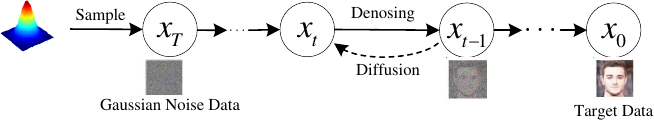}
		\label{fig1b}}
	\subfigure[Workflow of our MetaDiff]{
		\includegraphics[width=1.0\columnwidth]{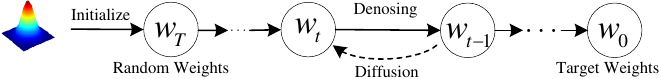}
		\label{fig1c}}
	\caption{Connection between gradient descent algorithm (GDA) and diffusion models. The gradient descent process (a) of GDA is similar to denoising process (b) of diffusion models. Based on this, we propose to model GDA as the denoising process of a diffusion model (c) and learn it in a diffusion manner, which does not need to differentiate through inner-loop path such that the issue of memory burdens and vanishing gradients can be alleviated for improving FSL. 
	}
	\vspace{-17pt}
	\label{fig1}
\end{figure}

To address this FSL problem, meta-learning \cite{rusu2018meta} has been proposed, which constructs a large number of base tasks from these base classes and then leverages it to learn task-agnostic meta knowledge for assisting novel class learning. Among these methods, gradient-based meta-learning methods are gaining increased attention, with its potential on generalization. This type of methods aims to learn a sample-efficient gradient descent algorithm (called meta-optimizer) by directly modeling its initialization \cite{finn2017model}, learning rate \cite{baik2020meta}, or update rule \cite{RaviL17} as a shared meta parameter and then learning it in a bi-level (\emph{i.e.}, outer-loop and inner-loop) optimization manner. Here, the outer-loop process accounts for learning a task-agnostic meta parameter for meta-optimizer, while the inner-loop process leverages the meta-optimizer to learn a task-specific base learner with few gradient updates. Although these methods have shown superior performance, the outer-loop process requires backpropagation along inner-loop optimization path and calculating second-order derivatives for learning meta-optimizer such that the significant memory overhead and risk of vanishing gradient is imposed during training \cite{rajeswaran2019meta}. This degrades meta-learning performance \cite{nichol2018first}. Some works attempt to address this issue but from the perspective of gradient approximate estimations \cite{rajeswaran2019meta, nichol2018first}, which would introduce an estimation error into the gradient (\emph{i.e.}, meta-gradient) for outer-loop optimization, and hamper its generalization ability. 


Inspired by recent diffusion models \cite{ho2020denoising}, we also focus on gradient-based meta-learning with its good generalization but present a new diffusion perspective to model gradient descent algorithm. It does not need to differentiate through inner-loop, such that the above issues of memory burden   and vanishing gradients can be alleviated for improving meta-learning. Specifically, as shown in Figures~\ref{fig1a} and \ref{fig1b}, we find that 1) the optimizaton process of gradient descent algorithm (see Figure~\ref{fig1a}) is very similar to the denoising process of diffusion models from a Gaussion noise to a target variable (see Figure~\ref{fig1b}). The key difference is that the denoising variable is model weight in gradient descent algorithm but origin image data in diffusion models; And 2) the latter is actually a generalized and learnable version of the former with weight momentum update and uncertainty estimation (see Section~\ref{sec_4_1}). In other words, the optimization process of gradient descent algorithm can be described as: given a randomly initial weight, the target weight is finally obtained by gradually removing its noise. 

Based on this fact, we propose a novel meta-learning with conditional diffusion, called MetaDiff. As shown in Figure~\ref{fig1c}, our idea is regarding the weight of base learner as a denoising variable, and then modeling the gradient descent algorithm as a diffusion model and learning it in a diffusion manner. Its key challenge of achieveing the above idea is how to predict the diffused noise of model weights at each time step $t$ with few labeled samples for a base learner. To address this challenge, we take few labeled samples as the condition of diffusion models and carefully design a gradient-based task-conditional UNet for noise prediction. Different from previous gradient-based meta-learning methods that learn the meta-optimizer in a bi-level optimization manner, our MetaDiff learns it in a diffusion manner. Thanks to its training efficiency and robustness, more superior meta-learning performance can be achieved for improving FSL.

Our main contributions can be summarized as follows:
\begin{itemize}
	\item We are the first to reveal the close connection between gradient descent algorithm and diffusion models. From workflow, we find that the optimization process of gradient descent algorithm is very similar to the denoising process of diffusion models. After theoretical analysis, the denoising process of diffusion models is actually a generalized and learnable gradient descent algorithm with weight momentum updates and uncertainty estimation. 
	
	\item Based on this fact, we propose a novel diffusion-based meta-learning for FSL. 
	In particular, a gradient-based conditional UNet is designed as our meta-learner for noise prediction. Thanks to diffusion training efficiency, the issue of memory burden and vanishing gradients can be effectively alleviated for improving meta-learning.
	
	\item We conduct comprehensive experiments on two public data sets, which verify the effectivenss of our MetaDiff.
\end{itemize}

\section{Related Work}
\label{sec:relatedwork}

\subsection{Meta-Learning}
Few-shot learning (FSL) is a challenging task, which aims to recognize novel classes with few examples \cite{chen2021meta, ChenLKWH19}. To address this problem, meta-learning is proposed, which aims to learn to quickly learn novel tasks with few examples \cite{flennerhag2019meta, zhu2023transductive}. The core idea is learning task-agnostic meta knowledge from a large number of similar tasks and then leveraging it to assist the learning of novel tasks. From the type of meta-knowledge, these existing meta-learning methods can be roughly grouped into three groups. \emph{The metric-based methods} \cite{snell2017prototypical, vinyals2016matching, zhang2021prototype, zhang2022sgmnet, zhang2022metanode, zhang2022hyperbolic, zhang2023prototype} regard the metric space or metric strategy as meta knowledge and perform the novel class prediction in a nearest-prototype manner. 
\emph{The model-based methods} \cite{hou2019cross, zhmoginov2022hypertransformer, li2019lgm} regard a black-box model as meta knowledge, which leverages it and few data to directly predict model weights or test sample labels. 
\emph{The gradient-based methods} \cite{rusu2018meta, lee2019meta, rajeswaran2019meta, nichol2018first, von2021learning, RaghuRBV20, zhang2023scalable} regard the gradient-based optimization algorithm as meta knowledge, which learn to model its hyperparameters (\emph{i.e.}, learning rate \cite{baik2020meta}, loss function \cite{baik2021meta}, initialization \cite{finn2017model}, preconditioner \cite{kang2023meta}, or updata rules \cite{deleu2022continuous, RaviL17}) such that the base learner can be quickly learned with few gradient updates. 


We focus on the gradient-based meta-learning due its good generalization. However, different from existing methods, we presents a new diffusion perspective to model meta-optimizer, which does not need to differentiate through inner-loop path such that the issues of memory burdens and vanishing gradients can be alleviated for improving FSL. 


\subsection{Diffusion Models}
Diffusion model \cite{ho2020denoising, nichol2021improved} is a popular type of deep generative models, which models and learns the generation process of target data from random Gaussion noises in a forward diffusion and reverse denoising manner. 
With the superior properties of diffusion models, 
the diffusion models have been widely exploited on various vision \cite{lugmayr2022repaint} and multi-modal tasks \cite{rombach2022high, kawar2023imagic, kumari2023multi} and achieved remarkable performance improvement. 
However, there is very few works (\emph{i.e.}, \cite{roy2022diffalign} and \cite{hu2023meta}) to explore diffusion models for FSL. Specifically, in \cite{roy2022diffalign}, Roy et al. introduce class names as priors and then leverage it and a text2image diffusion model to generate more images for alleviating the data-scarcity issue of FSL. Instead of using text2image diffusion models, Hu et al. \cite{hu2023meta} employ a image2image diffusion model and leverage it to generate more high-similarity pseudo-data for improving FSL.

Different from existing methods that regarding the diffusion model as a component of data augmentation, we find that the gradient descent process is similar to the denoising process, thus we propose to model the gradient descent algorithm as a diffusion model. We note that a concurrent working with our MetaDiff is ProtoDiff \cite{du2023protodiff}. However, different from ProtoDiff that focuses on metric-based meta-learning (\emph{i.e.}, rectifying prototype bias), we target at gradient-based meta-learning, and first reveal the close connections between gradient descent algorithm and diffusion models. Then, a new diffusion-based meta-optimizer is presented for fast adaptation of base-learner.

\section{Problem Definition and Preliminaries}
\subsection{Problem Definition}
For a $N$-way $K$-shot FSL problem, it consists of two datasets, \emph{i.e.}, a base class dataset $\mathcal{D}_{base}$ and a novel class dataset $\mathcal{D}_{novel}$. The base class dataset $\mathcal{D}_{base}$ consists of abundant labeled data from base class $\mathcal{C}_{base}$, which is used for assisting the classifier learning of novel classes. The novel class dataset $\mathcal{D}_{novel}$ contains two sub datasets from novel classes $\mathcal{C}_{novel}$, \emph{i.e.}, a training set (call support set $\mathcal{S}$) that consists of $N$ classes and $K$ samples per class and a test set (call query set $\mathcal{Q}$) consisting of unlabeled samples.

Our goal is that leveraging the base class dataset $\mathcal{D}_{base}$ to learn a good meta-optimizer such that the classifier can be quickly learned from few labeled data (\emph{i.e.}, the support set $\mathcal{S}$) to perform the novel class prediction for query set $\mathcal{Q}$.

\subsection{Preliminaries}
\label{sec_3_2}
\noindent{\bf Diffusion Models.} Diffusion models aim to model a probability transformation from a prior Gaussian distribution $p_{prior} \in \mathcal{N}(\mathbf{0}, \mathbf{I})$ to a target distribution $p_{target}$. 
It consists of two processes, \emph{i.e.}, a diffusion (also called forward) process and a denoising (also called reverse) process. 

\emph{1) The diffusion process} aims to iteratively add a noise from a Gaussian distribution to a target data $x_{0} \sim p_{target}$ to transform $x_{0}$ into $x_{1},x_{2}, ..., x_{T}$. The final $x_{T}$ tends to become a sample point from the prior distribution $p_{prior} \in \mathcal{N}(\mathbf{0}, \mathbf{I})$ when the number of iterations $T$ tends to big enough. The diffusion process aims to learn a noise prediction model $\epsilon_{\theta}(x_{t}, t)$ for estimating the added noise at time $t-1$ from $x_{t}$, which is then used to recovery the target data in denoising process. The training object $L$ is as follows:
\begin{equation}
	\begin{aligned}
		L=\mathbb{E}_{x_{0} \sim p_{target}, \epsilon \sim \mathcal{N}(\mathbf{0}, \mathbf{I}), t} [\|\epsilon-\epsilon_{\theta}(x_{t}, t)\|_2^2],
	\end{aligned}
	\label{eq_1}
\end{equation}where $\|\cdot\|_2^2$ denotes a mean squared error loss. It is worth noting that the above training object defined in Eq.~\eqref{eq_1} can be performed at any time step $t$ without the iterations of adding noise due to its good closed form at any time step $t$. That is,
\begin{equation}
	\begin{aligned}
		q(x_{t}|x{0}) &= \mathcal{N}(x_{t};\sqrt{\overline{\alpha}_{t}}x_{0}, (1-\overline{\alpha}_{t})\mathbf{I}), \\ &\alpha_{t}=1-\beta_t, \overline{\alpha}_{t}=\prod_{s=1}^{t}\alpha_{t},
	\end{aligned}
	\label{eq_2}
\end{equation}where $\beta_t \in (0,1)$ is a variance hyperparameter. 

\emph{2) The denoising process} is reverse process of diffusion. Based on the learned noise prediction model $\epsilon_{\theta}(x_{t}, t)$, given a start noise $x_T \sim p_{prior}$, we can iteratively remove its fraction of noises at each time $t$ and finally recovery the target data $x_{0}$ from the noisy data $x_T \sim \mathcal{N}(\mathbf{0}, \mathbf{I})$. That is, 
\begin{equation}
	\begin{aligned}
		x_{t-1}=\frac{1}{\sqrt{\alpha_{t}}}(x_{t}-&\frac{\beta_t}{\sqrt{(1-\overline{\alpha}_{t})}}\epsilon_{\theta}(x_{t}, t)) + \sigma_{t}z, z \sim \mathcal{N}(\mathbf{0}, \mathbf{I}).
	\end{aligned}
	\label{eq_4}
\end{equation}where $\sigma_{t}$ is a variance hyperparameter, which is theoretically set to $\sigma^2_{t}=\beta_t$ in most existing diffusion works \cite{ho2020denoising, nichol2021improved}.

\noindent{\bf Gradient Descent Algorithm (GDA).} GDA is a family of optimization algorithm, which aims to optimize model weights by following the opposite direction of gradient. Formally, let $w$ denotes the weights of a base learner $g_{w}(\cdot)$ and $L(w)$ be its differentiable loss function, and $\nabla L(w)$ be its weight gradient, during performing gradient descent algorithm. The overall optimization process of GDA can be summaried as iteratively performing Eq.~\eqref{eq_5}, that is,
\begin{equation}
	\begin{aligned}
		w_{t+1} = w_{t} - \eta (\nabla L(w_t)), t=0,1,,,T-1.
	\end{aligned}
	\label{eq_5}
\end{equation}where $w_0$ is an initial weight, \emph{i.e.}, a Gaussian noise in origin gradient descent algorithm; and $\eta$ denotes a learning rate. 
Due to the data scarcity issue in FSL, directly employing the Eq.~\eqref{eq_5} to learn a base learner $g_{w}(\cdot)$ would result in an overfitting issue. To address this issue, gradient-based meta-learning attempts to learn a GDA in a bi-level optimization (\emph{i.e.}, outer-loop and inner-loop) manner by modeling its hyperparameters (\emph{e.g.}, initial weight $w_0$ or learning rate $\eta$) as meta-knowledge for improving FSL. However, some studies \cite{rajeswaran2019meta} show the outer-loop process requires backpropagation along inner-loop optimization path such that the risk of vanishing gradient is imposed, which degrades meta-learning performance. In this paper, we propose a diffusion perspective to address this issue. 

\begin{figure*}
	\centering
	\includegraphics[width=1.0\textwidth]{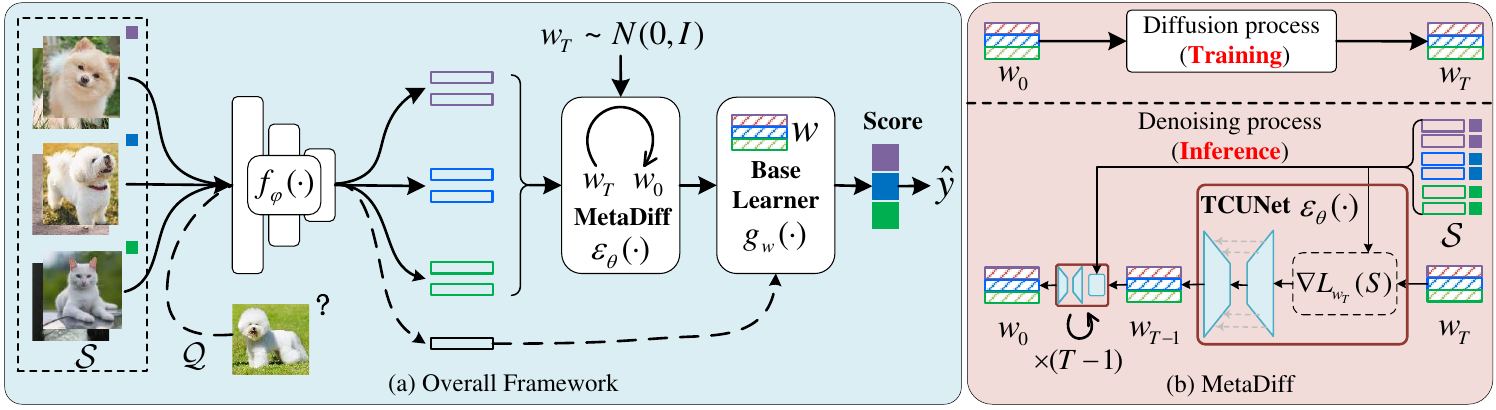}
	\caption{(a) The overall framework of our MetaDiff-based FSL method. (b) Illustration of our MetaDiff meta-optimizer $\epsilon_{\theta}(\cdot)$.  
	}
	\vspace{-15pt}
	\label{fig2}
\end{figure*}

\section{Methodology}
\subsection{Connection: Diffusion Models vs GDA}
\label{sec_4_1}

As introduced in Section~\ref{sec_3_2}, we can describe the process of denoising and gradient descent process as follows: 1) given a noise data $x_{T}$, the denoising process iteratively performs Eq.~\eqref{eq_4} to obtain a latent sequence $x_{T-1}, x_{T-2},...,x_{0}$. As a result, a target data $x_{0}$ can be recoveried from a Gaussian noise $x_{T}$; and 2) given a randomly initial wight $w_{0}$, the gradient descent process of GDA iteratively performs Eq.~\eqref{eq_5} to a latent sequence $w_{1}, w_{2},...,w_{T}$. As a result, an optimizal weight $w_{T}$ can be obtained from a random weight $w_{0}$. We can see that the denoising process of diffusion models and gradient descent process of GDA are very similar in workflow. This inspires us to think about what is the close connection between the two methods in theory. Let's take a closer look on Eqs.~\eqref{eq_4} and ~\eqref{eq_5}, Eq.~\eqref{eq_4} first can be simplifed as:
\begin{equation}
	\begin{aligned}
		x_{t-1}=\underbrace{\frac{1}{\sqrt{\alpha_{t}}}}_{Term 1}x_{t}-&\underbrace{\frac{\beta_t}{\sqrt{\alpha_{t}}\sqrt{(1-\overline{\alpha}_{t})}}}_{Term 2}\epsilon_{\theta}(x_{t}, t) + \underbrace{\sigma_{t}}_{Term 3}z,\\ z \sim &\mathcal{N}(\mathbf{0}, \mathbf{I}).
	\end{aligned}
	\label{eq_6}
\end{equation}Let $\gamma$ denotes the \emph{Term 1}, $\eta$ be the \emph{Term 2}, $\xi$ be the \emph{Term 3} of Eq.~\eqref{eq_6}, respectively. The Eq.~\eqref{eq_6} can be simplifed as:
\begin{equation}
	\begin{aligned}
		x_{t-1}=\gamma x_{t}-\eta \epsilon_{\theta}(x_{t}, t) + \xi z,\ z \sim \mathcal{N}(\mathbf{0}, \mathbf{I}).
	\end{aligned}
	\label{eq_7}
\end{equation}Due to $\gamma>1$ ($\alpha_{t}<1$), we can transform Eq.~\eqref{eq_7} as follows:
\begin{equation}
	\begin{aligned}
		x_{t-1}=\underbrace{x_{t}-\eta \epsilon_{\theta}(x_{t}, t)}_{Term 1}& + \underbrace{(\gamma - 1) x_{t}}_{Term 2}+\underbrace{\xi z}_{Term 3},\\ z \sim \mathcal{N}&(\mathbf{0}, \mathbf{I}).
	\end{aligned}
	\label{eq_7_}
\end{equation}where $Term 1$ is denoising term, $Term 2$ denotes a momentum update term with hyperparameter $\gamma-1$ (also called exponentially weighted moving average), and $Term 3$ is a uncertain term. Comparing Eqs.~\eqref{eq_5} and \eqref{eq_7_}, we can see that the gradient decent process defined in Eq.~\eqref{eq_5} is equivalent to the $Term 1$ of Eq.~\eqref{eq_7_}, which means that Eq.~\eqref{eq_5} is a special case of denoising process described in Eq.~\eqref{eq_7_} when the $\gamma$ is set to one (\emph{i.e.}, $\gamma=1$),  the $\eta$ is regarded as a hyperparameter, and the $\xi$ is set to zero. In particular, it is worth noting that the predicted variable of noise prediction model $\epsilon_{\theta}(x_{t}, t)$ is actually the gradient (\emph{i.e.}, $\nabla L(w)$) of model weights. In other words, the denosing process defined in Eq.\eqref{eq_7_} can be viewed as a generalized and learnable gradient descent algorithm defined in Eq.\eqref{eq_5}, \emph{i.e.}, a learnable gradent descent algorithm with weight momentum updates and uncertainty estimation where $\gamma$ controls the weight of momentum updates, $\eta$ is a learning rate, and $\xi$ is the degree of uncertainty.

\emph{Why set parameters ($\gamma$, $\eta$, and $\xi$) by following Eq.~\ref{eq_6}?} Instead of using manual setting or model learning manner like existing meta-optimizers to set hyperparameters $\gamma$, $\eta$, and $\xi$, respectively, the diffusion models unify the parameter settings by theoretical deduction, \emph{i.e.}, $\gamma=\frac{1}{\sqrt{\alpha_{t}}}$, $\eta=\frac{\beta_t}{\sqrt{\alpha_{t}}\sqrt{(1-\overline{\alpha}_{t})}}$, and $\xi=\sigma_{t}$ where $\alpha_{t}=1-\beta_t$, $\sigma^2_{t}=\beta_t$, and $\beta_t$ is experimentally set in linear decreasing manner from a small value (e.g., $10^{-4}$) to a large value (e.g., 0.02). The goal of such setting is to ensure that denoising and diffusion processes have approximately the same functional form and the efficiency and robustness of diffusion training (\emph{i.e.}, the training objective defined in Eq.~\eqref{eq_1} can be performed at any time step $t$ without the iterations from $t=0$ to $t$).

\subsection{Meta-Learning with Conditional Diffusion}
Inspired by the above analysis, we find that the diffusion model is a generalized and learnable form of GDA and its hyperparameter setting have rigorous theoretical derivation, which enables its inspiring advantage (\emph{i.e.}, generation robustness and training efficiency). Based on this, we attempt to leverage a diffusion model to model GDA and then present a new meta-optimizer, \emph{i.e.}, MetaDiff, for fast adaptation of base-learner. It does not need to differentiate through inner-loop path, such that the memory burden and risk of vanishing gradients can be alleviated for improving FSL.

\noindent{\bf Overall Framework.} The overall framework of our MetaDiff on FSL is presented in Figure~\ref{fig2}(a), which consists of an embedding network $f_{\varphi}(\cdot)$ with parameters $\varphi$, a base learner $g_{w}(\cdot)$ with parameters $w$, and a MetaDiff meta-optimizer $\epsilon_{\theta}(\cdot)$ with meta parameters $\theta$. Here, the embedding network $f_{\varphi}(\cdot)$ aims to encode each support/query image as a $d$-dim feature vector. Inspired by prior meta-learning works \cite{deleu2022continuous, lee2019meta}, we assume that the embedding network $f_{\varphi}(\cdot)$ is shared across tasks, which can be obtained by using a simple pretraining manner on entire base class classification task \cite{chen2021meta, ChenLKWH19}. The base learner $g_{w}(\cdot)$ is a simple linear or prototype classifer (the prototype classifer is used in this paper due it good performance), which is a task-specific and needs to be adapted starting at some Gaussian initialization $w_{T}$. The MetaDiff $\epsilon_{\theta}(\cdot)$ is a meta-optimizer, which takes the features and labels of all support samples $(u_i, y_i) \in \mathcal{S}$ as inputs and then learns a target weights $w_{0}$ for base learner $g_{w}(\cdot)$ from initial weights $w_{T}$ in a denoising manner (see Figure~\ref{fig2}(b)).

Specifically, given a $N$-way $K$-shot FSL task, we first leverage the embedding network $f_{\varphi}(\cdot)$ to encode the feature $f_{\varphi}(u_i)$ for each support/query image $u_i \in \mathcal{S} \cup \mathcal{Q}$. Then, we randomly  initialize a weight $w_{T} \sim \mathbb{N}(\mathbf{0}, \mathbf{I})$ for the base learner $g_{w}(\cdot)$, and design a task-conditional UNet (\emph{i.e.}, the noise prediction model $\epsilon_{\theta}(\cdot)$) that regards the features and labels of all support sample $(u_i, y_i) \in \mathcal{S}$ as task condition, to estimate the noise to be removed at time $t$. After that, we take the weight $w_{T}$ as the denoising variable and iteratively perform the denoising process from $t=T$ to $t=1$, that is,
\begin{equation}
	\begin{aligned}
		w_{t-1}=\frac{1}{\sqrt{\alpha_{t}}}(w_{t}-&\frac{\beta_t}{\sqrt{(1-\overline{\alpha}_{t})}}\epsilon_{\theta}(w_{t}, \mathcal{S}, t)).
	\end{aligned}
	\label{eq_8}
\end{equation}Note that we remove the uncertainty term (\emph{i.e.}, $\sigma_{t}z$) for deterministic estimation during inference. After iteratively perform $T$ step, the target weight $w_0$ can be obtained as the optimal weight $w$ for base learner $g_{w}(\cdot)$. Finally, we perform class prediction of each query image $u_i \in \mathcal{Q}$ by using the learned optimal base learner $g_{w}(\cdot)$. That is,
\begin{equation}
	\begin{aligned}
		\hat{y}=g_{w}(f_{\varphi}(u_i)),\ w=w_{0},\ u_i \in \mathcal{Q}.
	\end{aligned}
	\label{eq_9}
\end{equation}Here, we only introduce the inference workflow of our MetaDiff-based FSL framework, which is summaried in Algorithm~\ref{alg:sampling}. Next, we will introduce the design details of our key component, \emph{i.e.}, the task-conditional UNet $\epsilon_{\theta}(\cdot)$.

\begin{figure}
	\centering
	\includegraphics[width=1.0\columnwidth]{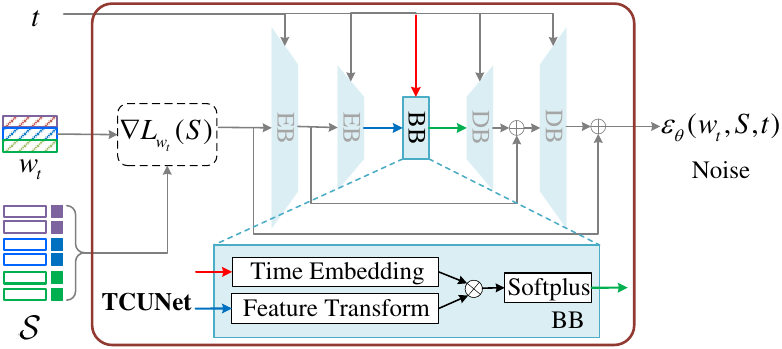}
	\caption{Illustration of our task-conditional UNet (\emph{i.e.}, TCUNet). ``EB'', ``BB'', and ``DB'' denotes the encoder, bottle, and decoder blocks, repsectively. The details of ``EB'', ``BB'', and ``DB'' are all similar. For clarity, we only show the design details of ``BB'' in figure and others are similar.}
	\vspace{-15pt}
	\label{fig3}
\end{figure}

\noindent{\bf Task-Conditional UNet (TCUNet).} The task-conditional UNet (TCUNet) $\epsilon_{\theta}(\cdot)$ is the key component in our MetaDiff meta-optimizer, which takes the features and labels of all support samples $(u_i, y_i) \in \mathcal{S}$, time step $t$, and the weight $w_t$ of base learner as inputs. It aims to estimate the noise to be remove for the weight $w_t$ of base learner at each time step $t$. We attempt to use a general conditional UNet like \cite{rombach2022high} for implementing TCUNet $\epsilon_{\theta}(\cdot)$. However, we find that such general conditional UNet does not work in our MetaDiff, which inspires us to think deeply the rationale of the noise prediction model $\epsilon_{\theta}(\cdot)$ in our MetaDiff meta-optimizer. As analyzed in Section~\ref{sec_4_1}, we can see that the goal of noise prediction model $\epsilon_{\theta}(\cdot)$ is actually equivalent to predict gradient in meta-optimizer (see Eqs.~\eqref{eq_5} and \eqref{eq_7_}). 

\algrenewcommand\algorithmicindent{0.5em}%
\begin{figure}[t]
	\begin{minipage}[t]{0.470\textwidth}
		\begin{algorithm}[H]
			\caption{Training} \label{alg:training}
			\small
			\begin{algorithmic}[1]
				\Repeat
				\State Sampling a task $\tau=(\mathcal{S}, \mathcal{D}_{base}^{\tau})$ from datasets $\mathcal{D}_{base}$
				\State Estimating $w_{0}$ by training $g_{w}(\cdot)$ $\ $ on auxiliary datasets $\mathcal{D}_{base}^{\tau}$
				\State Sampling time $t \sim \mathrm{Uniform}(\{1, \dotsc, T\})$
				\State Sampling $\epsilon \sim \mathcal{N}(\mathbf{0}, \mathbf{I})$
				\State Take gradient descent on $\nabla_{\theta} \|\epsilon-\epsilon_{\theta}(w_{t}, \mathcal{S}, t)\|_2^2$, \emph{i.e.}, Eq.~\ref{eq_11}
				\Until{converged}
			\end{algorithmic}
		\end{algorithm}
	\end{minipage}
	\hfill
	\begin{minipage}[t]{0.470\textwidth}
		\begin{algorithm}[H]
			\caption{Inference} \label{alg:sampling}
			\small
			\begin{algorithmic}[1]
				\State Given a $N$-way $K$-shot task $\tau=(\mathcal{S}, \mathcal{Q})$ from novel classes
				\State Sampling a random weight $w_{T} \sim \mathcal{N}(\mathbf{0}, \mathbf{I})$
				\For{$t=T, \dotsc, 1$}
				\State Performing Eq.~\ref{eq_8}	$w_{t-1}=\frac{1}{\sqrt{\alpha_{t}}}(w_{t}-\frac{\beta_t}{\sqrt{(1-\overline{\alpha}_{t})}}\epsilon_{\theta}(w_{t}, \mathcal{S}, t))$
				\EndFor
				\State Performing class prediction of query samples by Eq.~\ref{eq_9}
			\end{algorithmic}
		\end{algorithm}
	\end{minipage}
	\vspace{-15pt}
\end{figure}

Based on this find, as shown in Figure \ref{fig3}, we design a task-conditional UNet from the perspective of gradient estimation as the noise prediction model $\epsilon_{\theta}(\cdot)$. The key idea is predicting noise from the view of gradient estimation instead of a general black-box manner like \cite{rombach2022high}.  Specifically, given a base learner weight $w_t$ at time $t$ and all features and labels of support samples $(u_i, y_i) \in \mathcal{S}$, we first leverage the base learner $g_{w_t}(\cdot)$ with model weights $w_t$ to compute the loss $L_{w_t}(\mathcal{S})$ of of all support samples. That is,
\begin{equation}
	\begin{aligned}
		L_{w_t}(\mathcal{S})=\frac{1}{|\mathcal{S}|} \sum_{(u_i, y_i) \in \mathcal{S}} loss\_fun(g_{w_t}(f_{\phi}(u_i)), y_i).
	\end{aligned}
	\label{eq_10}
\end{equation}where $|\cdot|$ is the number of support samples and $loss\_fun(\cdot)$ is a loss function. Instead of cross-entropy loss, we employ a simple L2 loss to implement the $loss\_fun(\cdot)$, that is,
\begin{equation}
	\begin{aligned}
		L_{w_t}(\mathcal{S})=\frac{1}{|\mathcal{S}|} \sum_{(u_i, y_i) \in \mathcal{S}} \|w_{t,y_i}-f_{\phi}(u_i)\|_2^2,
	\end{aligned}
	\label{eq_10_}
\end{equation}where $w_{t,y_i} \in w_t$ is the class prototype of label $y_i$. The intuition of such design is moving the class prototype towards the center of all labeled sample for each class $y_i$, which is more matching for the rationale of prototype classifier. Then, the gradient $\nabla L_{w_t}(\mathcal{S})$ regarding weights $w_t$ can be obtained as the initial noise estimation for the base learner $g_{w_t}(\cdot)$ at time $t$. 
To obtain more accuracy noise estimation, we take the initial noise estimation $\nabla L_{w_t}(\mathcal{S})$ as inputs and then design a conditional UNet fusing time embedding $t$ to predict the noise to be remove at time $t$ for base learner $g_{w}(\cdot)$. 

\begin{table*}
	\begin{center}
		\smallskip\scalebox
		{0.84}{
			\begin{tabular}{l|c|c|c|c|c|c}
				\hline
				\multicolumn{1}{l|}{\multirow{2}{*}{Method}}&\multicolumn{1}{c|}{\multirow{2}{*}{Adaptation Type}}&\multicolumn{1}{c|}{\multirow{2}{*}{Backbone}}& \multicolumn{2}{c|}{miniImagenet} & \multicolumn{2}{c}{tieredImagenet} \\ 
				\cline{4-7}
				& & & 5-way 1-shot & 5-way 5-shot & 5-way 1-shot & 5-way 5-shot \\
				\hline
				\hline
				iMAML\cite{rajeswaran2019meta} & All & Conv4 & $49.30 \pm 1.88\%$  & $59.77 \pm 0.73\%$ & $38.54 \pm 1.37\%$  & $60.24 \pm 0.76\%$ \\
				ALFA \cite{baik2020meta} & All & Conv4 & $50.58 \pm 0.51\%$  & $69.12 \pm 0.47\%$ & $53.16 \pm 0.49\%$  & $70.54 \pm 0.46\%$ \\
				MeTAL \cite{baik2021meta} & All & Conv4 & $52.63 \pm 0.37\%$  & $70.52 \pm 0.29\%$ & $54.34 \pm 0.31\%$  & $70.40 \pm 0.21\%$ \\
				MetaSGD + SiMT \cite{tack2022meta} & All & Conv4 & $51.70 \pm 0.80\%$  & $69.13 \pm 1.40\%$ & $52.98 \pm 0.07\%$  & $71.46 \pm 0.12\%$ \\
				GAP \cite{kang2023meta} & All & Conv4 & $54.86 \pm 0.85\%$  & $71.55 \pm 0.61\%$ & $57.60 \pm 0.93\%$  & $74.90 \pm 0.68\%$ \\
				ANIL \cite{RaghuRBV20} & only CH & Conv4 & $46.30 \pm 0.40\%$  & $61.00 \pm 0.60\%$ & $49.35 \pm 0.26\%$  & $65.82 \pm 0.12\%$ \\
				COMLN \cite{deleu2022continuous} & only CH & Conv4 & $53.01 \pm 0.62\%$  & $70.54 \pm 0.54\%$ & $54.30 \pm 0.69\%$  & $71.35 \pm 0.57\%$ \\ 
				MetaQDA \cite{zhang2021shallow} & only CH & Conv4 & $\textbf{56.41} \pm \textbf{0.80}\%$  & $72.64 \pm 0.62\%$ & $\textbf{58.11} \pm \textbf{0.48}\%$  & $74.28 \pm 0.73\%$ \\ 
				\hline
				MetaDiff (ours) & only CH & Conv4 & 55.06 $\pm$ 0.81$\%$ & \textbf{73.18} $\pm$ \textbf{0.64}$\%$ & 57.77 $\pm$ 0.90$\%$ & \textbf{75.46} $\pm$ \textbf{0.69}$\%$  \\	
				\hline \hline
				
				ALFA \cite{baik2020meta} & All & ResNet12 & $59.74 \pm 0.49\%$  & $77.96 \pm 0.41\%$ & $64.62 \pm 0.49\%$  & $82.48 \pm 0.38\%$ \\
				MAML+SiMT \cite{TackPLLS22} & All & ResNet12 & 62.05 $\pm$ 0.39$\%$ & 78.77 $\pm$ 0.45$\%$ & 63.91 $\pm$ 0.32$\%$ & 77.43 $\pm$ 0.47$\%$ \\
				LEO \cite{rusu2018meta} & only CH & WRN-28-10 & $61.76 \pm 0.08\%$  & $77.59 \pm 0.12\%$ & $66.33 \pm 0.05\%$  & $81.44 \pm 0.09\%$ \\
				Meta-Curvature\cite{park2019meta} & only CH & WRN-28-10 &  $61.85 \pm 0.10\%$  & $77.02 \pm 0.11\%$ & $67.21 \pm 0.10\%$  & $82.61 \pm 0.08\%$ \\
				ANIL\cite{RaghuRBV20} & only CH & ResNet12 &  $49.65 \pm 0.65\%$  & $59.51 \pm 0.56\%$ & $54.77 \pm 0.76\%$  & $69.28 \pm 0.67\%$ \\
				COMLN \cite{deleu2022continuous} & only CH & ResNet12 & 59.26 $\pm$ 0.65$\%$ & 77.26 $\pm$ 0.49$\%$ & 62.93 $\pm$ 0.71$\%$ & 81.13 $\pm$ 0.53$\%$ \\
				ClassifierBaseline \cite{chen2021meta} & only CH & ResNet12 & 61.22 $\pm$ 0.84$\%$ & 78.72 $\pm$ 0.60$\%$ & 69.71 $\pm$ 0.88$\%$ & 83.87 $\pm$ 0.64$\%$ \\
				MetaQDA \cite{zhang2021shallow} & only CH & ResNet18 & \textbf{65.12} $\pm$ \textbf{0.66}$\%$ & 80.98 $\pm$ 0.75$\%$ & 69.97 $\pm$ 0.52$\%$ & 85.51 $\pm$ 0.58$\%$ \\
				\hline
				MetaDiff (ours) & only CH & ResNet12 & 64.99 $\pm$ 0.77$\%$ & \textbf{81.21} $\pm$ \textbf{0.56}$\%$ & \textbf{72.33} $\pm$ \textbf{0.92}$\%$ & \textbf{86.31} $\pm$ \textbf{0.62}$\%$ \\	
				\hline			
		\end{tabular}}
	\end{center}
	\caption{Experiment results on ImageNet derivatives. The best results are highlighted in bold. ``CH'' denotes classification head. }
	\vspace{-15pt}
	\label{table1}
\end{table*}

As shown in Figure \ref{fig3}, the UNet consists of two encoder blocks (EB), a bottle block (BB) and two decoder blocks (DB). At each encoder step, we halve the number of input features and then remain unchanged at bottle step, but the number of features is doubled at each decoder step. The details of each encoder, bottle, decoder block are all similar, which contains a feature transform layer, a time embedding layer, and a ReLU activation layer. Note that we remove the ReLU activation layer in the final decoder block for estimating gradients. At each block, its output is obtained by first feeding the output of previous block and time step $t$ into the feature transform and time embedding layers, respectively, and then fusing them in an element-by-element product manner, finally followed by a softplus activation.

\noindent{\bf Meta-Learning Objective.}  Different from previous gradient based meta-learning methods that learn a meta-optimizer in a bi-level optimization manner, as shown in Figure~\ref{fig2}(b), we employ a diffusion process to train our MetaDiff. However, unlike existing diffusion models  \cite{ho2020denoising}  where the target data $x_0$ is known (\emph{i.e.}, origin images), the target variable of our MetaDiff is model weight (\emph{i.e.}, $w_{0}$) of base learner $g_{w}(\cdot)$ which is unknown. Thus, a key challenge of training our MetaDiff is how to obtain a large number of target weight $w_{0}$ for base learner $g_{w}(\cdot)$. 

To this end, we follow episodic training strategy \cite{vinyals2016matching} and construct a large number of $N$-way $K$-shot tasks from base class dataset $\mathcal{D}_{base}$. Then, given a constructed $N$-way $K$-shot tasks $\tau$, based on its origin label $k'$ of each class $k=0,1,..,N-1$ in the base classes $\mathcal{C}_{base}$, we extract all samples that belongs to the origin label $k'$ of each class $k=0,1,..,N-1$ from the base class dataset $\mathcal{D}_{base}$, as the auxiliary dataset $\mathcal{D}_{base}^{\tau}$. The labeled data is very sufficient in the auxiliary dataset $\mathcal{D}_{base}^{\tau}$ because it contains all labeled data belonging to class $k'$ in $\mathcal{D}_{base}$, thus we can leverage it to learn a base learner $g_{w}(\cdot)$ such that the target weight $w_0$ can be obtained for each task $\tau$. Finally, we leverage the target weight $w_0$ of all constructed tasks to train our MetaDiff meta-optimizer in a diffusion manner. That is,
\begin{equation}
	\begin{aligned}
		\mathop{min}_{\theta} \mathbb{E}_{(\mathcal{S}, w_0) \sim \mathbb{T}, \epsilon \sim \mathcal{N}(\mathbf{0}, \mathbf{I}), t \sim [1,T]} \|\epsilon-\epsilon_{\theta}(w_{t}, \mathcal{S}, t)\|_2^2.
	\end{aligned}
	\label{eq_11}
\end{equation}During training, our MetaDiff does not require backpropagation along the inner-loop optimization path and calculating second-order derivatives for learning meta-optimizer such that the memory overhead and the risk of vanishing gradient can be effectively alleviated for improving FSL.  The complete diffusion procedure is summaried in Algorithm~\ref{alg:training}. 

\section{Experiments}
\subsection{Datasets and Settings}
\noindent \textbf{MiniImagenet.} It is a subset from ImageNet, which contains 100 classes and 600 images per class. Following \cite{lee2019meta}, we split it into three sets, \emph{i.e.}, 64, 16, and 20 classes for training, validation, and test, respectively. 

\noindent \textbf{TieredImagenet.}
It is also a ImageNet subset but larger, which has 608 classes and 1200 images per class. Following \cite{lee2019meta}, it is splited into 20, 6, and 8 high-level classes for training, validation, and test, respectively. 

\begin{table}[!t]
	\centering
	\smallskip
	\smallskip\scalebox
	{0.76}{
		\smallskip\begin{tabular}{c|l|c|c}
			\hline
			& Method &  5-way 1-shot & 5-way 5-shot \\
			\hline \hline
			(\romannumeral1) & Baseline (Standard GDA) & 60.53 $\pm$ 0.86$\%$ & $72.43 \pm 0.66\%$  \\
			(\romannumeral2) & Replacing by Momentum GDA & 62.03 $\pm$ 0.82$\%$ & 78.28 $\pm$ 0.56$\%$ \\
			(\romannumeral3) & Replacing by ANIL & 60.77 $\pm$ 0.82$\%$ & 77.34 $\pm$ 0.64$\%$ \\
			(\romannumeral4) & Replacing by MetaLSTM & 63.56 $\pm$ 0.81$\%$ & 79.90 $\pm$ 0.59$\%$\\
			(\romannumeral5) & Replacing by ALFA & 63.92 $\pm$ 0.82$\%$ &80.01 $\pm$ 0.61$\%$ \\
			(\romannumeral6) & Replacing by Our MetaDiff & 64.99 $\pm$ 0.77$\%$ & $81.21 \pm 0.56\%$ \\
			\hline
	\end{tabular}}
	\caption{Analysis of our MetaDiff on miniImagenet.
	}
    \vspace{-15pt}
	\label{table3}
\end{table}

\begin{table}[!t]
	\centering
	\smallskip
	\smallskip\scalebox
	{0.87}{
		\smallskip\begin{tabular}{c|c|c|c}
			\hline
			& Method & 5-way 1-shot & 5-way 5-shot \\
			\hline \hline
			(\romannumeral1) & TCUNet & 64.99 $\pm$ 0.77$\%$ & 81.21 $\pm$ 0.56$\%$ \\
			(\romannumeral2) & Replacing L2 loss & 62.92 $\pm$ 0.79$\%$ & 80.92 $\pm$ 0.56$\%$ \\
			(\romannumeral3) & w/o UNet  & 62.72 $\pm$ 0.84$\%$ & 80.72 $\pm$ 0.55$\%$ \\
			\hline
	\end{tabular}}
	\caption{Analysis of our TCUNet on miniImagenet.}
	\vspace{-15pt}
	\label{table4}
\end{table}

\subsection{Implementation Details}

\noindent \textbf{Network Details.}
We use Conv4 and ResNet12 as the embedding network $f_{\phi}(\cdot)$, which are same to previous gradient-based meta-learning methods \cite{kang2023meta, lee2019meta, deleu2022continuous} and delivers 128/512-dim vector for each image. In our task-conditional UNet, 
for encoder blocks, we use a linear layer with 512/256-dim inputs and 256/128-dim outputs to implement its feature transform layer, and a linear layer with 32-dim inputs and 256/128-dim outputs as its time embedding layer. For bottle blocks, we use a linear layer with 128-dim inputs and outputs to implement its feature transform layer, and a linear layer with 32-dim inputs and 128-dim outputs as its time embedding layer. For decoder blocks, we use a linear layer with 128/256-dim inputs and 256/512-dim outputs to implement its feature transform layer, and a linear layer with 32-dim inputs and 256/512-dim outputs as its time embedding layer.

\noindent \textbf{Training Details.}
During training, we train our MetaDiff meta-optimizer 30 epochs (10000 iterations per epoch) using Adam with a learning rate of 0.0001 and a weight decay of 0.0005. Following the standard setting of diffusion models in \cite{ho2020denoising}, we set the number of denoising iterations to 1000 (\emph{i.e.}, $T=1000$ is used).


\begin{table}[!t]
	\centering
	\smallskip
	\smallskip\scalebox
	{0.77}{
		\smallskip\begin{tabular}{c|l|c|c}
			\hline
			& Method & 5-way 1-shot & 5-way 5-shot \\
			\hline \hline
			\multicolumn{1}{l|}{\multirow{2}{*}{(\romannumeral1)}} & Prototype Classfier (ALFA) & 63.92 $\pm$ 0.82$\%$ & 80.01 $\pm$ 0.61$\%$ \\
			& Prototype Classfier (MetaDiff) & 64.99 $\pm$ 0.77$\%$ & 81.21 $\pm$ 0.56$\%$ \\
			\hline
			\multicolumn{1}{l|}{\multirow{2}{*}{(\romannumeral2)}} & Linear Classfier (ALFA) & 62.09 $\pm$ 0.84$\%$ & 78.13 $\pm$ 0.59$\%$ \\
			& Linear Classfier (MetaDiff) & 62.72 $\pm$ 0.89$\%$ & 80.19 $\pm$ 0.57$\%$ \\
			\hline
	\end{tabular}}
	\caption{Classifier analysis of MetaDiff on miniImagenet.}
	\vspace{-10pt}
	\label{table5}
\end{table}

\subsection{Experimental Results}
Our MetaDiff falls into the type of gradient-based meta-learning, thus we mainly select various state-of-the-art gradient-based meta learning methods as our baselines. We evaluate our MetaDiff and these baselines on Imagenet derivatives. The experimental results are shown in Table~\ref{table1}. Among them, iMAML\cite{rajeswaran2019meta}, MAML \cite{finn2017model}, ALFA \cite{baik2020meta}, ANIL \cite{RaghuRBV20}, COMLN \cite{deleu2022continuous}, GAP \cite{kang2023meta}, LEO \cite{deleu2022continuous}, and ClassifierBaseline \cite{chen2021meta}, are our key competitors, which also focus on learning GDA but modeling its hyperparameters. 

Table~\ref{table1} shows the results of various gradient-based meta-learning methods on miniImagenet and tieredImagenet. From these results, we find that (\romannumeral1) our MetaDiff achieves superior or comparable performance on all tasks, which exceeds most state-of-the-art gradient-based meta-learning by around 1\% $\sim$ 3\%. This verifies the effectiveness of our MetaDiff; and (\romannumeral2) Our MetaDiff achieves consistent improvement on Conv4 and ResNet12 backbones for all tasks, which is reasonable because our MetaDiff mainly focuses on the adaptation of classification head. This also verifies the universality of our MetaDiff on various backbones. 


\begin{figure}[t]
	\centering
	\subfigure[MetaLSTM vs MetaDiff]{ 
		\includegraphics[width=0.45\columnwidth]{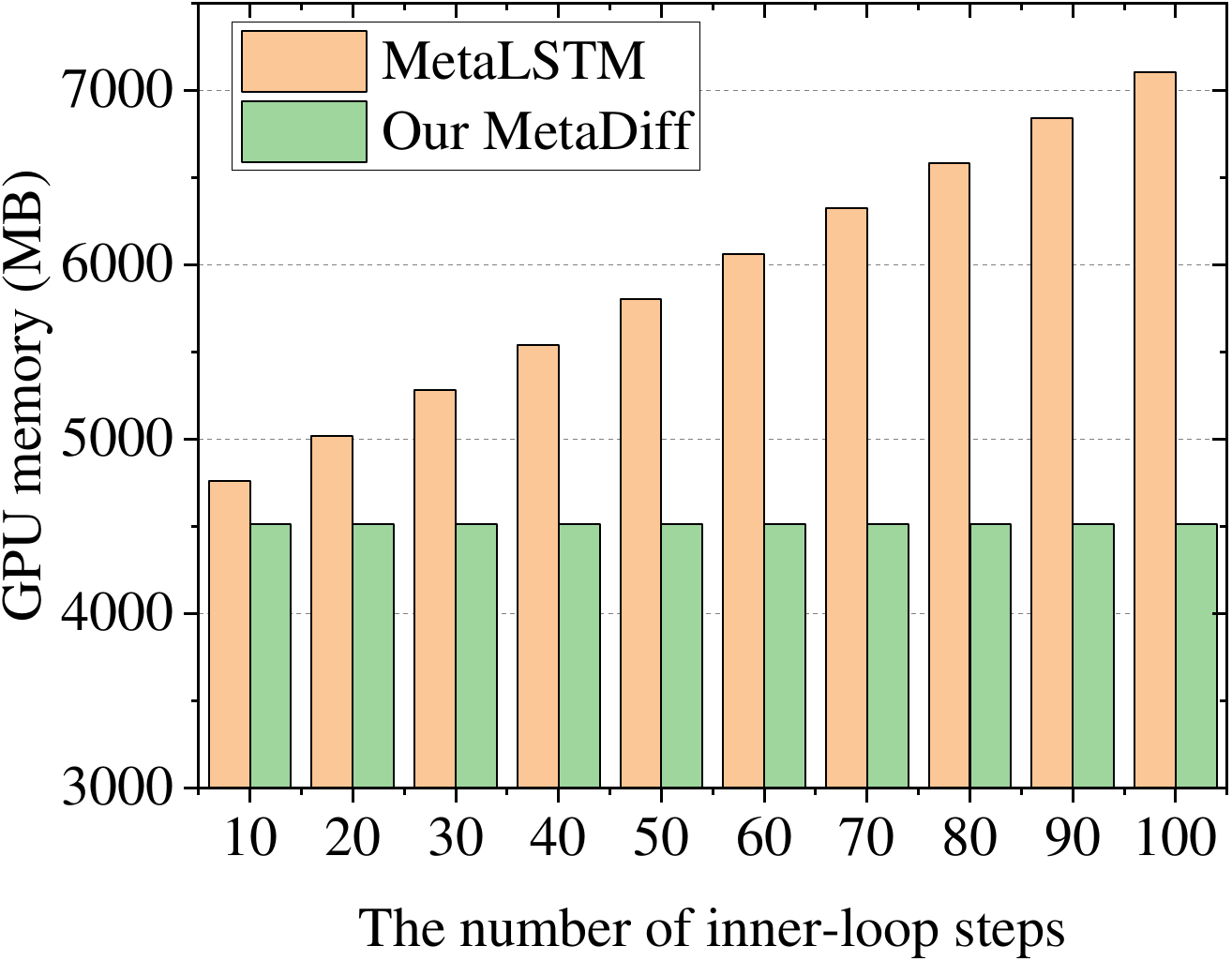}
		\label{fig4a}}
	\quad
	\subfigure[ALFA vs MetaDiff]{ 
		\includegraphics[width=0.45\columnwidth]{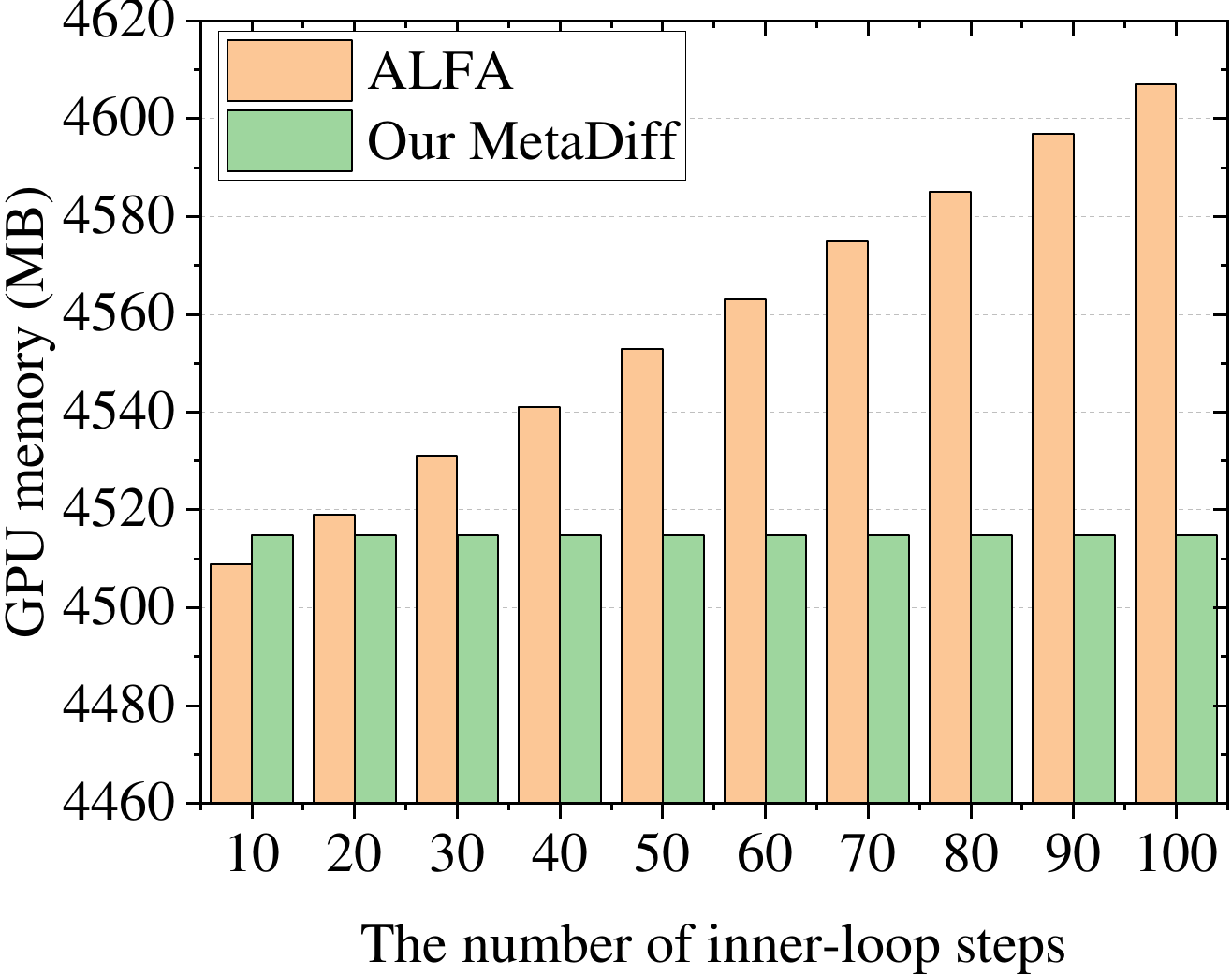}
		\label{fig4b}}
	\caption{GPU memory on 1-shot tasks of miniImagenet. 
	}
	\vspace{-10pt}
	\label{fig4}
\end{figure}

\begin{figure}[t]
	\centering
	\subfigure[Test Accuracy Curve]{ 
		\includegraphics[width=0.45\columnwidth]{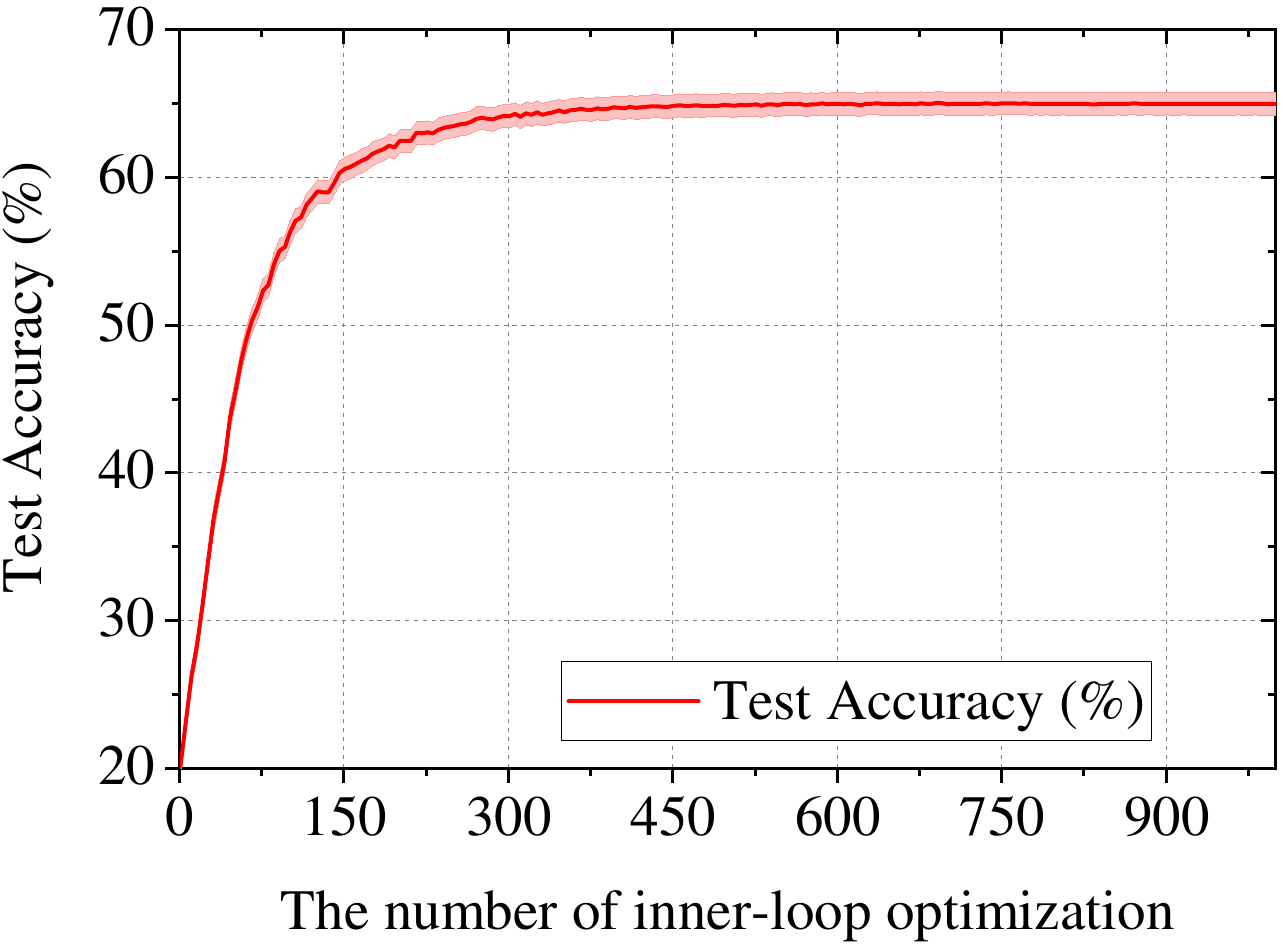}
		\label{fig5a}}
	\quad
	\subfigure[Test Loss Curve]{ 
		\includegraphics[width=0.45\columnwidth]{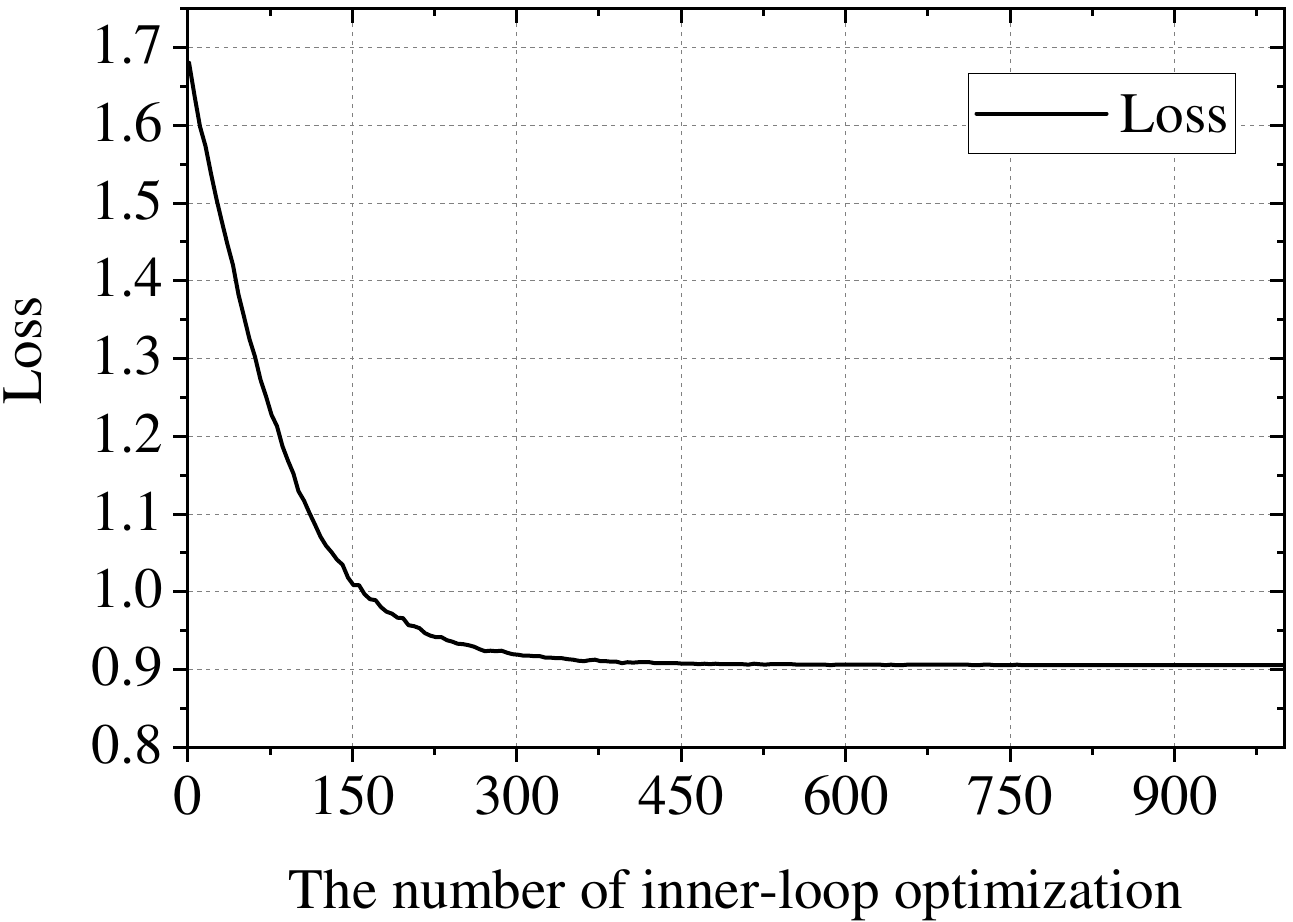}
		\label{fig5b}}
	\caption{Convergence Analysis on  miniImagenet. 
	}
	\vspace{-15pt}
	\label{fig5}
\end{figure}

\subsection{Ablation Study}
\label{section_5_4}

\noindent {\bf Is our MetaDiff effective?} In Table~\ref{table3}, we analyze the effectiveness of our MetaDiff. Specifically, (\romannumeral1) we implement the adaptation of base learner (\emph{i.e.}, the prototype classfier) by using a standard GDA (\emph{i.e.}, Eq.~\eqref{eq_5}) on the support set $\mathcal{S}$; (\romannumeral2) we replace the standard GDA (\emph{i.e.}, Eq.~\eqref{eq_5}) by a GDA with gradient momentum updates on (\romannumeral1); (\romannumeral3) replacing by the ANIL \cite{RaghuRBV20} on (\romannumeral1); (\romannumeral4) replacing by the MetaLSTM \cite{RaviL17} on (\romannumeral1); (\romannumeral5) replacing by the ALFA \cite{baik2020meta} on (\romannumeral1); and (\romannumeral6) replacing by our MetaDiff. From the experimental results of (\romannumeral1) $\sim$ (\romannumeral7), we observe that: 1) the performance of (\romannumeral2) $\sim$ (\romannumeral6) exceeds (\romannumeral1) around 1\% $\sim$ 5\%, which means that it is helpful to learn a meta-optimizer to optimize task-specific base-learner; 2) the performance of (\romannumeral7) exceeds (\romannumeral2) $\sim$ (\romannumeral6) around 1\% $\sim$ 4\%, which shows the superiority of our MetaDiff. 


\noindent {\bf Are our task-conditional UNet effective?} In Table~\ref{table4}, (\romannumeral1) we evaluate TCUNet on miniImagenet; (\romannumeral2) we replace the L2 loss defined in Eq.~\eqref{eq_10_} by using cross-entropy loss; (\romannumeral3) we remove the UNet on (\romannumeral1). From results, we can see that the performance of our TCUNet descreases by around 1\% $\sim$ 3\% when removing UNet or replacing L2 loss by cross-entropy loss. This implies that leveraging the idea of gradient-based UNet and L2 loss to estimate noise is useful for our TCUNet. 


\noindent {\bf Can our MetaDiff be applied to other classifiers?} To verify the universality of our MetaDiff on other classifiers, in Table~\ref{table5}, we evaluate our MetaDiff and ALFA on prototype classifiers and linear classifiers. We find that our MetaDiff all achieves superior performacne on these two classifier and prototype classifier performs better. This result implies that our MetaDiff is very universal for different classifiers.

%


\subsection{Statistical Analysis}
\label{section_5_5}

\noindent {\bf How much our MetaDiff take GPU memory?} In Figure~\ref{fig4}, we select MetaLSTM \cite{RaviL17} and ALFA \cite{baik2020meta} as baselines and report the GPU memory during training by varing the number of inner-loop number. From Figure~\ref{fig4}, we can see that 1) the cost of GPU memory keep increase linearly as the number of inner-loop step increase; however 2) our MetaDiff keep constant. This is reasonable because our MetaDiff is trained in a diffusion manner, which is irrelevant to inner-loop optimization.

\noindent {\bf Can our MetaDiff converge?} We randomly select 600 5-way 1-shot tasks from the test set of miniImageNet, and then report their test accuracy and loss of entire denoising process. The results are shown in Figure~\ref{fig5}. From the result, we can observe that our MetaDiff can converge to a stable result within a finite number of steps, around 450 steps.


\section{Conclusion}
In this paper, we present a novel meta-learning with conditional diffusion for few-shot learning, called MetaDiff. In particular, we find that the diffusion model actually is a generalized version of gradient descent, a learnable gradient descent algorithm with weight momentum updates and uncertainty estimation, and then design a task-conditional UNet from the perspective of gradient estimation to predict the denoising nosie for target weights. Experimental results on two public data sets verify the effectiveness of our MetaDiff. 

\section*{Acknowledgments}
This work was supported by the NSFC under Grant No. 62272130 and Grant No. 62376072, and the Shenzhen Science and Technology Program under Grant No. KCXFZ20211020163403005.

\bibliography{aaai24}

\end{document}